\pdfoutput=1

\documentclass[11pt]{article}

\usepackage{ACL2023}

\usepackage{times}
\usepackage{latexsym}
\usepackage{amssymb}
\usepackage{booktabs}
\usepackage{amsfonts}
\usepackage{caption}
\usepackage{threeparttable}
\usepackage{geometry}
\usepackage{multirow}
\usepackage{microtype}
\usepackage{booktabs}
\usepackage{algorithm}
\usepackage[noend]{algpseudocode}
\usepackage{subcaption}
\usepackage{amsmath}
\usepackage{graphicx}
\usepackage{multicol}
\usepackage{xspace}
\usepackage[T1]{fontenc}
\usepackage{pifont}
\newcommand{\xmark}{\ding{55}}%
\newcommand{\templama}{\textsc{Temp\-LAMA}\xspace}
\newcommand{\tempreason}{\textsc{Temp\-Reason}\xspace}

\usepackage[utf8]{inputenc}

\usepackage{microtype}
\definecolor{gred}{RGB}{255,102,102}
\definecolor{gblue}{RGB}{51,102,255}
\definecolor{gyellow}{RGB}{244,180,0}
\definecolor{ggreen}{RGB}{15,157,88}
\definecolor{ggrey}{RGB}{115,115,115}
\definecolor{na}{gray}{0.9}
\definecolor{LightYellow}{RGB}{255,255,191}
\definecolor{OrangeRed}{rgb}{1.0, 0.27, 0.0}
\definecolor{midnightgreen}{rgb}{0.0, 0.29, 0.33}
\definecolor{darkgreen}{rgb}{0.0, 0.42, 0.24}
\definecolor{skyblue}{RGB}{70, 130, 180}
\usepackage{colortbl}
\newcommand*{\affmark}[1][*]{\textsuperscript{#1}}
\makeatletter
\def\thanks#1{\protected@xdef\@thanks{\@thanks
        \protect\footnotetext{#1}}}
\makeatother
\usepackage{tablefootnote}

\usepackage{inconsolata}

\title{Towards Benchmarking and Improving the Temporal Reasoning Capability of Large Language Models}

\author{Qingyu Tan\thanks{$^{*}$Qingyu Tan is under the Joint PhD Program between Alibaba and NUS.} \affmark[$^{*}$ 1, 2]~~~\textbf{Hwee Tou Ng\affmark[$^\dag$ 2] \thanks{$^\dag$  Corresponding author.}~~~Lidong Bing\affmark[1] } 
\\$^1$DAMO Academy, Alibaba Group~~\\
$^2$Department of Computer Science, National University of Singapore\\
\texttt{\{qingyu.tan,l.bing\}@alibaba-inc.com}\\
\texttt{\{qtan6,nght\}@comp.nus.edu.sg}\\
}

\begin{document}
\maketitle
\begin{abstract}
Reasoning about time is of fundamental importance. Many facts are time-dependent. For example, athletes change teams from time to time, and different government officials are elected periodically. Previous time-dependent question answering (QA) datasets tend to be biased in either their coverage of time spans or question types. In this paper, we introduce a comprehensive probing dataset \tempreason to evaluate the temporal reasoning capability of large language models. Our dataset includes questions of three temporal reasoning levels. In addition, we also propose a novel learning framework to improve the temporal reasoning capability of large language models, based on temporal span extraction and time-sensitive reinforcement learning. We conducted experiments in closed book QA, open book QA, and reasoning QA settings and demonstrated the effectiveness of our approach\footnote{Our code and data are released on \url{https://github.com/DAMO-NLP-SG/TempReason}}. 
\end{abstract}

\begin{figure*}
    \centering
    \resizebox{\textwidth}{!}{
    \includegraphics{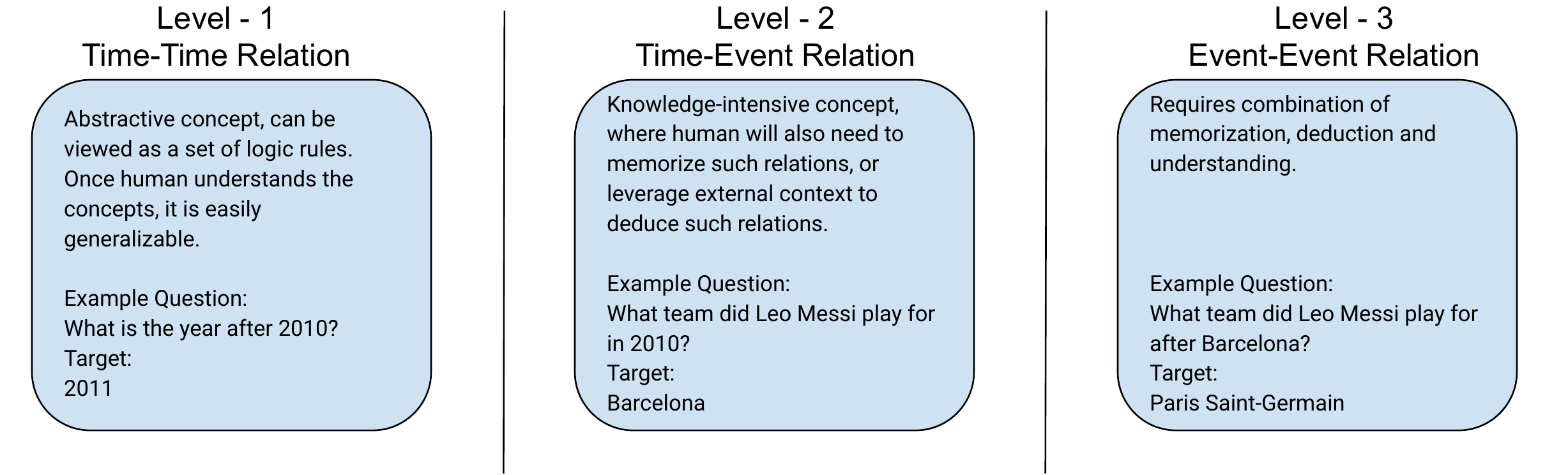}
    }
    \setlength{\belowcaptionskip}{-10pt}
    \caption{Illustration of three levels of understanding towards time. }
    \label{fig:understanding-fig}
\end{figure*}

\section{Introduction}

In recent years, large language models (LLMs) have achieved significant success in many natural language processing (NLP) tasks, such as natural language understanding (NLU) \cite{fei-etal-2023-isa}, information extraction (IE) \cite{ding-etal-2023-annotator}, and question answering (QA) \cite{ye-etal-2023,zhao-etal-2023-verify}. Many facts and answers are dependent on their related time scopes, such as `What soccer club was Lionel Messi playing for?'. \citet{chia2022dataset} has pointed out around 48\% of the qualifiers in the widely-used knowledge base Wikidata~\citep{vrandevcic2014wikidata} are time-related. That is, a significant number of the knowledge triples in the Wikidata KB have their expiry dates. Correct understanding of temporal concepts is crucial for language models to be successful in real-world applications. To examine the temporal reasoning capabilities of LLMs, the Time-Sensitive Question Answering (TSQA) task has been proposed and several evaluation datasets were published for research purposes. The Time-sensitive QA dataset \cite{chen2021dataset} and the \templama dataset \citep{10.1162/tacl_a_00459} were constructed based on the Wikidata temporal KB. StreamingQA~\citep{liska2022streamingqa} was constructed by news article collections in English WMT challenges from 2007 to 2020. One consensus of prior work is that time-sensitive QA is a challenging task and its performance is still far below human performance. However, they did not provide a systematic analysis of LM's temporal reasoning capability. In this paper, we aim to systematically analyze such capability and identify the strengths and weaknesses of LMs on temporal reasoning.

As shown in Figure~\ref{fig:understanding-fig}, humans' understanding of temporal reasoning could be broken down into three levels: time-time (L1) relation, time-event (L2) relation, and event-event (L3) relation. For the understanding of time-time relations, humans can easily determine the relation between two timestamps $t_1$ and $t_2$ on the time axis. For example, when humans are asked `What is the year after 2020?', they are able to answer this question without any external information. This level of temporal understanding could be regarded as a set of logic rules and is highly generalizable across different times, while this type of reasoning was overlooked in prior TSQA research (\citealp{ning-etal-2020-torque}; \citealp{chen2021dataset}; \citealp{10.1162/tacl_a_00459}). For time-event relations, the reasoning process requires grounding events to their specific time ranges. In this paper, the concept of events includes time-dependent facts. Humans either memorize a large number of time-event pairs or need to rely on relevant contexts to deduce such relations. An example question is `What soccer club was Lionel Messi playing for in Dec 2010?', where a time is specified in the question, and the answer changes based on the given time. If this question is posed to a person who is unfamiliar with sports, this person also needs external information to provide the answer. Answering this type of questions requires information retrieval and temporal grounding. For event-event relations, there are multiple reasoning paths to determine such relations. One possible path is to first identify the timestamps of different events and perform time-time reasoning. Another path is to search for the textual cues of relative relation, such as `before', `after', `during', and `simultaneous'.

 We first conducted a simple preliminary experiment for probing LLM's L1 temporal reasoning capability. We found that not only do LMs perform poorly on the time-time relation task, but they are also heavily biased in favor of contemporary years (2000 - 2020). This may be due to the imbalanced term frequencies in the pre-training corpora. Most LLMs (such as BERT, GPT, and T5) are pre-trained on raw texts from a snapshot at a specific timestamp, typically around 2018 to 2020. Therefore, the time expression vocabulary is highly dependent on term frequencies in the pre-training corpora. Typically, year tokens that occur frequently will have a smaller index in the vocabulary and the uncommon years generally have larger indices or will be split into subtokens. Take the T5 tokenizer as an example, the year `2014' is tokenized as `2014', however, the year `2021' is tokenized as `20' and `21'. This means that language models only learn the co-occurrences of time expressions and their context. 
 
Given such findings, we found that the recently proposed TSQA \templama dataset has several main drawbacks. Firstly, the time span of the dataset is only from 2010 to 2020, which is a highly biased distribution in favor of LM. Secondly, it only focused on the questions of time-event relations. To overcome these shortcomings, we created a more comprehensive TSQA benchmark \tempreason, which spans a longer time range and all three types of temporal understanding. We conducted comprehensive experiments in closed book QA, open book QA, and reasoning QA settings. We found that the temporal reasoning capabilities of LLMs are highly variable with respect to the reference time in the question. LLMs perform well on the contemporary years and poorly on low-resource years.

Moreover, we proposed a novel temporal learning framework based on temporal span extraction and time-sensitive reinforcement learning. Our proposed framework encourages LMs to generate temporally correct answers while penalizing predictions that do not satisfy the temporal constraints. Experimental results showed that our proposed benchmark \tempreason provides a more comprehensive evaluation for LM's temporal reasoning capability and our model consistently outperforms strong baselines. 


\begin{table}[ht]
\centering
\resizebox{\columnwidth}{!}{
\begin{tabular}{lll} 
\hline
 Ref. Year    & Question                      & Target  \\
\hline
2011 & What is the year x years before 2011? & 2011 - x    \\
2010 & What is the year before 2010? & 2009    \\
1949 & What is the year x years after 1949? & 1949 + x    \\
1905 & What is the year after 1905? & 1906    \\

\hline
\end{tabular}}
\setlength{\belowcaptionskip}{-10pt}
\caption{Templates used for year prediction (yearly level). The reference year and interval x are randomly generated, where the reference year is within a specified time range and x $\leq$ 10. All the answers to this question are numeric representations of years.} 
\label{tab:templates-prelim}
\end{table}

\section{Preliminaries}
\label{sec:prelim}
We aim to examine the capability of LMs for simple year prediction. We first design a set of question templates that reflects the basic concepts of temporal prediction, as shown in Table~\ref{tab:templates-prelim}. Questions of these kinds can be easily answered by humans and this understanding is highly generalizable across the years, and all the expected answers are years in numeric form. In order to have a more comprehensive understanding of temporal expressions, we divide 1900 to 2040 into seven 20-year time periods. Then, we randomly generate 400 questions for each 20-year time period. We then use three language models to make predictions on such questions. The first LM is T5-large model fine-tuned on the Natural Question dataset (T5-L-NQ, \citealp{kwiatkowski2019natural}). This QA dataset is one of the largest open domain QA datasets. \citet{roberts-etal-2020-much} has demonstrated that language models fine-tuned on such data can achieve competitive performance on the open domain QA task. The second LM is FLAN-T5-Large \citep{wei2022finetuned} model. This model is instruction-tuned on data of more than 60 NLP tasks. The fine-tuned model demonstrated competitive zero-shot reasoning capability, and achieved strong performance on many natural language understanding and generation tasks. The third model is the popular ChatGPT \citep{ouyang2022training} model. To ensure that the predictions are consistent, we used the \textit{gpt-3.5-0301} version of ChatGPT.  We aim to evaluate the temporal reasoning capability of the three language models. We evaluate the answers using the following three metrics: (1) exact match (EM), which is a standard metric for QA. Besides, since the expected answers are numeric, we also evaluate the answers by (2) mean absolute error (MAE) and (3) trend accuracy (Trend Acc). Trend accuracy is calculated by whether the predicted year is before or after the reference year. If the trend is correct, the prediction is deemed to be correct.  

The experimental results on year prediction are shown in Table~\ref{tab:prelim-exp}. We report the scores of T5-L-NQ on the left, FLAN-T5-L in the middle, and ChatGPT on the right.
From these experiments, we have several interesting observations: (1) The ChatGPT model is able to solve this problem with high accuracy (99.6 overall EM). However, it still made a few mistakes in the 1900-1940 time period. (2) The first two LMs (T5-L-NQ and FLAN-T5-L) are \textbf{biased towards contemporary time ranges}. We can clearly see that the EM scores between 2000 to 2020 are significantly higher than the rest of the time ranges. This could be the result of the higher term frequencies of the contemporary year tokens in the pre-training corpora. Since many large LMs are trained and released after 2018, the pre-training corpora may contain more year expressions that are closer to that date. In contrast, the first two LMs perform significantly worse in the past (1900-2000) and the future (2020-2040) years. (3) The first two LMs \textbf{lack numeric reasoning ability} with respect to time. The answers provided by these LMs for the time prediction questions are in numeric form, indicating that the LMs understand what the questions are asking. However, the EM scores are all around the 20-30 range, except for T5-L-NQ in the 2000-2020 time range. This indicates that LMs have poor estimation of temporal concepts. Besides, we find that the FLAN-T5-L model has significantly lower EM scores compared to T5-L-NQ, but achieves lower MAE estimations across most of the time ranges. This indicates that instruction tuning implemented in FLAN has implicitly improved the numeric reasoning capability of T5. (4) On the other hand, \textbf{All LMs are good at catching (before/after) trends}, indicating that at least the LMs understand the concepts of before/after well. We can see that all LMs achieve over 90\% performance across time ranges before 2020. However, for the first two LMs, this capability is not able to generalize to the future, as the performance in 2020-2040 is significantly worse than in other time periods.

\begin{table}[t!]
\centering
\resizebox{\columnwidth}{!}{
\begin{tabular}{lccc}
\hline
\textbf{Time Range}  & \multicolumn{1}{l}{\textbf{EM} (\textcolor[rgb]{0.416,0.659,0.31}{↑})} & \multicolumn{1}{l}{\textbf{MAE} (\textcolor[rgb]{0.416,0.659,0.31}{\textbf{↓}})} & \multicolumn{1}{l}{\textbf{Trend Acc} (\textcolor[rgb]{0.416,0.659,0.31}{↑})}  \\
\hline
1900-1920  &  17.5/6.8/99.5	                                                          & 28.0/7.4/\textbf{0.0}	                                                                & 99.5/96.8/\textbf{100}                                                               \\
1920-1940     & 31.5/1.8/98.9	                                                           & 16.4/11.9/0.1	                                                                      & 94.5/94.5/\textbf{100}                                                                   \\
1940-1960           & 17.5/3.3/\textbf{100} 	                                                           & \textbf{7.7}/9.2/\textbf{0.0}	                                                                     & \textbf{100}/91.0/\textbf{100}                                                                  \\
1960-1980  & 22.5/3.5/\textbf{100} 	                                                          & 17.1/7.5/\textbf{0.0}                                                                   & 94.0/92.0/\textbf{100}                                                                   \\
1980-2000  & 23.0/10.0/\textbf{100} 	                                                          & 7.9/6.9/\textbf{0.0}	                                                                     & 98.5/\textbf{100}/\textbf{100}                                                                 \\
2000-2020   & \textbf{47.5}/\textbf{20.0}/\textbf{100}	                                                          & 51.2/\textbf{2.3}/\textbf{0.0}                                                                    & 97.0/\textbf{100}/\textbf{100}                                                                  \\
2020-2040                        & 23.5/11.3/\textbf{100}	                                                          & 15.7/8.9/\textbf{0.0}	                                                                    & 84.5/83.8/\textbf{100}                  \\
Average                                                  & 26.1/8.1/99.6		                                                          & 20.6/7.7/\textbf{0.0}	                                                                    & 95.4/94.0/\textbf{100}                  \\
\hline
\end{tabular}
}
\setlength{\belowcaptionskip}{-10pt}
\caption{Evaluation results of T5-L-NQ\tablefootnote{\url{https://huggingface.co/google/t5-large-ssm-nq}}~\citep{raffel2020exploring} (left), FLAN-T5-large\tablefootnote{\url{https://huggingface.co/google/flan-t5-large}}~\citep{wei2022finetuned} (middle), and ChatGPT (right) models on the year prediction task across different time ranges. \textbf{Bold scores} refer to the best performance of each model in each column.}
\label{tab:prelim-exp}

\end{table}

\section{Comprehensive Benchmark for Temporal Reasoning}
\label{sec:benchmark}
\subsection{\tempreason Dataset}

Based on the findings of the previous section, we found that the recently proposed \templama TSQA dataset~\citep{10.1162/tacl_a_00459} has several major limitations. Firstly, it only contains questions from 2010 to 2020, which are highly in favor of LM's temporal reasoning biases. Secondly, the \templama dataset is heavily biased towards long-duration facts, as 70.69\% of the questions of \templama have the most frequent answer for a given subject. That is, the \templama dataset may encourage models to learn shortcuts to memorize the most frequent answers instead of learning temporal reasoning capability. If the research on time-sensitive question answering only focuses on adaptation to a short period of time, the maintenance and continual adaptation shall be highly expensive. As shown in the previous section, language models perform poorly on past and future time spans. If the language model is not able to understand the changes from the past to the present, it is highly difficult for this model to understand the evolution from the present to the future. 
\begin{figure*}
    \centering
    \resizebox{\textwidth}{!}{
    \includegraphics{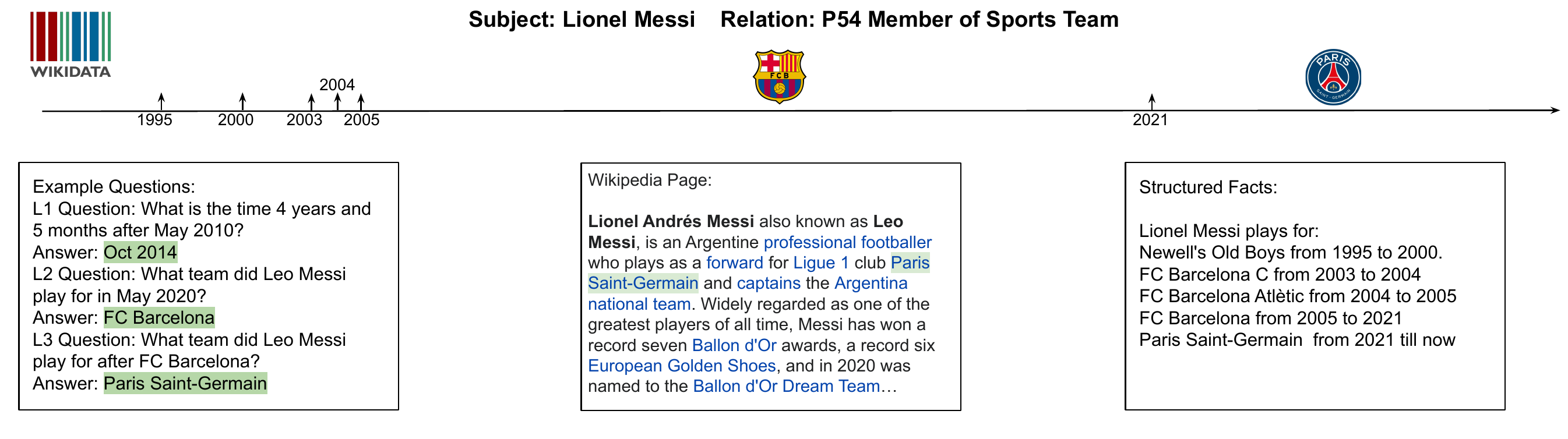}}
    \caption{Sample \tempreason questions and contexts. For humans, the L1 question can be answered without any context provided, whereas for L2 and L3 questions, humans will need to ground the events to timestamps and then perform temporal reasoning.}
    \label{fig:temp-r-example}
\end{figure*}
In order to probe the temporal reasoning ability in a more systematic manner, we constructed a new comprehensive dataset \tempreason. For the L1 time-time relation reasoning, we extend the year prediction task to month prediction, since year prediction can be enumerated by several thousands of examples and LMs may simply memorize such examples. Specifically, we randomly pick a reference time $t$ within a specific time range and then synthesize questions with respect to that time. The questions have the form of `What is the date x years and y months before/after $t$?'. In this way, we can randomly generate L1 questions and answers within the time period. To avoid data leakage, we make sure each generated question is unique. We then randomly split the questions to train, dev, and test sets. To evaluate the generalizability of L1 temporal reasoning, we also create a future test set from 2022 to 2040.

\begin{table}[t]
\centering

\resizebox{\columnwidth}{!}{
\begin{tabular}{lccc} 
\hline
\multicolumn{1}{l}{} & Train    & Dev     & Test     \\ 
\hline
Time Range           & 1014-2022 & 634-2023 & 998-2023  \\
L1-Questions         & 400,000    & 4,000    & 4,000     \\
L2-Questions         & 16,017    & 5,521    & 5,397     \\
L3-Questions         & 13,014    & 4,437    & 4,426     \\
Subjects             & 3,000     & 1,000    & 1,000     \\
Facts                & 16,017    & 5,521    & 5,397     \\
Facts/subjects        & 5.3      & 5.5     & 5.4      \\
\hline
\end{tabular}}
\setlength{\belowcaptionskip}{-10pt}
\caption{Dataset statistics of \tempreason. }
\label{tab:data-stats}
\end{table}
For L2 and L3 reasoning, similar to \citet{10.1162/tacl_a_00459} and \citet{chen2021dataset}, we also leverage the Wikidata KB as the knowledge source. We first preprocess the 20 Nov 2022 dump of the Wikidata \citep{vrandevcic2014wikidata} knowledge base (KB) to extract all time-dependent facts. We then keep the facts of 10 time-sensitive relations mentioned in the \templama dataset. We process the knowledge triples and qualifiers into quintuplet format, $(s, r, o, t_{s}, t_{e})$, where $s$ is the subject, $r$ is the relation, $o$ is the object, $t_{s}$ and $t_{e}$ are the start time and end time of this fact. We group all the temporal facts by $s$ and $r$. In this way, facts in the same group are all relevant to the subject $s$. The group of facts can be denoted as $S = \{(s, r, o_{i}, t_{s_{i}}, t_{e_{i}}) | i \in 1 ... N \}$ and they are sorted chronologically, where $N$ is the number of facts within a group. Since we mainly want to focus on questions whose answers change with time, we only keep the groups that contain three or more temporal facts. In this way, we make sure that each group has at least three time-dependent answers. Moreover, since the Wikidata KB is highly class-imbalanced, we only keep a maximum of 2,000 subjects for each relation type. We then create cloze-style questions based on time-dependent facts. For the time-event (L2) type of questions, we randomly select a time $t_{r}$ between $t_{s}$ and $t_{e}$, and we then create a question with the query $(s, r, ?, t_{r})$ and a set of manually-defined question templates. The templates can be found in Table~\ref{tab:question-templates} in Appendx~\ref{sec:question-templates}. For the event-event (L3) type of questions, we first identify the `before/after' relation pairs within group $S$ (we only keep the 1-hop pairs). We then create the event-event question for each `before/after' pair using similar templates of the L2 questions (Table~\ref{tab:question-templates}). The statistics of our \tempreason dataset can be found in Table~\ref{tab:data-stats}. We also compared our datasets with prior works in Appendix~\ref{sec:compare-dataset}

\subsection{Problem Settings}
\label{sec:prob-settings}
The time-sensitive question answering (TSQA) task is formally defined as follows: given an input question and its corresponding time (Figure~\ref{fig:temp-r-example}), the model is asked to output the answer of this question, and the answers are evaluated by token-level F1 and exact match (EM) scores. Intuitively, the difficulty of the TSQA task is highly dependent on the context provided for each question. The challenges of the TSQA task can be broken down into three levels: (1) \textbf{Answer Retrieval}. The first challenge of TSQA is finding the possible answers, which is the same challenge as normal open-domain question answering. For questions in \tempreason, each question may have 5.3 to 5.5 possible answers (Table~\ref{tab:data-stats}). (2) \textbf{Time Grounding}. The second challenge of TSQA is temporal grounding. That is, this sub-task is to find the start time and end time of each possible answer. (3) \textbf{Temporal Reasoning}. The last challenge is finding the correct answer among the possible candidates based on the specified time constraints. 

To thoroughly examine the temporal reasoning capability of large language models in different aspects, we propose to tackle TSQA in three different context settings: (1) closed book QA, (2) open book QA, and (3) reasoning QA. We describe the three problem settings as follows.

\noindent\textbf{Closed Book Question Answering (CBQA)}. CBQA is a common task formulation in time-sensitive QA research (\citealp{10.1162/tacl_a_00459}; \citealp{liska2022streamingqa}). In this setting, only the question is prompted to the language model, which is then asked to output the answer without access to any natural language text. In Figure \ref{fig:temp-r-example}, the example question is asking about the soccer athlete \textit{Lionel Messi}. The most difficult part of this question is the memorization of \textit{Lionel Messi}'s experiences, since people who are not sports fans may not be able to answer such questions easily. 

\noindent\textbf{Open Book Question Answering (OBQA)}. The OBQA formalization is a more realistic problem setting, where external context in the form of natural language text is provided to help LMs to answer the questions. As shown in middle of Figure~\ref{fig:temp-r-example}, we use the Wikipedia page of the subject entity as part of the prompt to the language model, together with the question. 

\noindent\textbf{Reasoning QA}. In this setting, all the relevant temporal facts within the group $S = \{(s, r, o_{i}, t_{s_{i}}, t_{e_{i}}) | i \in 1 ... N \}$ are provided in structured form as part of the prompt (right of Figure~\ref{fig:temp-r-example}). This is a simplified version of OBQA since all possible answers and their time ranges are provided in the context. To avoid the models learning shortcuts, the provided facts are re-ordered randomly. Essentially, this setting resembles human temporal reasoning. The language models are required to deduce answers based on the time ranges of all possible answers. Human is able to deduce the answer by locating the query time within the group. Intuitively, human-level performance in this setting can be regarded as 100\%.

\section{Improving Temporal Reasoning}
\label{sec:method}
In order to improve the temporal reasoning capabilities, we propose a temporal training framework for sequence-to-sequence language models. Firstly, we pre-train the language model with a temporal span extraction task to encourage the model to pay more attention to the temporal and entity spans. We then fine-tune the model on task-specific data in \tempreason. Finally, we further fine-tune the language model by time-sensitive reinforcement learning with our novel reward function.

\noindent\textbf{Temporal Span Extraction Pre-Training (TSE)} Conventional language model pre-training randomly masks texts and reconstructs the original sentence. However, the relative importance of tokens and spans differs. \citet{guu2020retrieval} first introduced salient span masking, i.e, reconstructing masked named entities, as an intermediate pre-training technique for language models. This approach has shown positive effects on the QA task. In order for the language model to capture more knowledge on time-related spans, we first pre-train on 100K Wikipedia articles with a temporal and entity span extraction task. Specifically, we use the Spacy NER tagger to extract the temporal and entity spans in 100K Wikipedia articles. The NER tagger is trained on the Ontonotes 5.0 corpus \citep{weischedel2013ontonotes}. We randomly mask 50\% of the entities and temporal spans for a given paragraph and treat this paragraph as the input of T5 models. In this way, the model pays more attention to the contexts that are relevant to temporal shifts. Then the pre-trained language model will be used for fine-tuning with \tempreason question-answer pairs in different settings.

\noindent\textbf{Supervised Fine-Tuning (SFT)} The TSE pre-trained language model with parameters $\theta$ will then be fine-tuned on the task data in each setting. The input prompt to the LM is the concatenation of question $q$ and context $c$, and the objective of SFT is to maximize the probability of $P(a|q,c)$, where $a$ is the correct answer.

\begin{figure}[ht]
    \centering
    \resizebox{0.8\columnwidth}{!}{
    \includegraphics{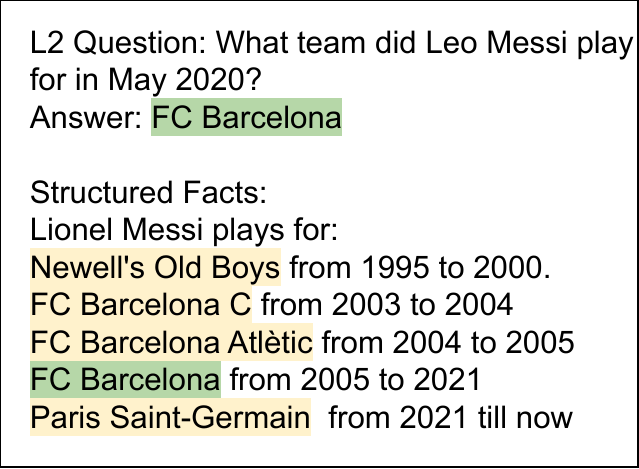}}
    \setlength{\belowcaptionskip}{-10pt}
    \caption{An example of time-sensitive reinforcement learning (TSRL). The ground truth is highlighted in green color and the negative answers are highlighted in yellow color.}
    \label{fig:rl-example}
\end{figure}

\noindent\textbf{Time-Sensitive Reinforcement Learning (TSRL)} One of the key challenges of temporal reasoning is that there are multiple possible answers for one subject. For a given fact $x = (s,r,o_{j}, t_{s_{j}}, t_{e_{j}})$, we have the facts in the same group $S_{N} = \{(s, r, o_{i}, t_{s_{i}}, t_{e_{i}}) | i \in 1 ... N, i\neq j \}$. These facts have the same subject and relation as the given fact, but are in other time periods. Therefore, for a question related to the fact $x$, we are able to collect the negative answers $N = \{o_{i} | i \in 1 ... N, i\neq j \}$ within the same group as the negative sample set for TSQA. An example of such negative examples is shown in Figure~\ref{fig:rl-example}. For a given question related to fact $x$, we want to maximize the probability of the correct answer $o_{j}$ while penalizing the model when it outputs temporally wrong answers. The correct answers and negative answers were used for our reward function. We first calculate the positive score $p(x)$ of the model prediction $\theta(x)$ with respect to the ground truth:
\begin{equation}
    p(x) = F(\theta(x),o_{j})
\end{equation}
where $F$ refers to the scoring function for reward computation. Specifically, we used the $EM$ scoring function as $F$.  We then calculate the negative score $n(x)$ by:
\begin{equation}
    n(x) = max\{F(\theta(x),o_{i})|i \neq j \}
\end{equation}
The negative score will be 1 if the model prediction returns a temporally wrong answer. Finally, the reward function for TSRL is calculated as: 
\begin{equation}
 R(x)=\left\{
\begin{array}{rcl}
p(x)&      & p(x) \geq n(x)\\
- n(x)   &      & n(x) > p(x)\\
\end{array} \right.
\end{equation}
The reward function is designed to give positive rewards for predictions that match the ground truth and negative rewards for predictions that match the answers in the negative answer set $N$.
We then optimize the fine-tuned language model by the Proximal Policy Optimization \citep{schulman2017proximal} algorithm. We denote our final model as TempT5.

\section{Experiments}
\subsection{Experimental Settings}
We conduct experiments in each proposed setting in Section \ref{sec:prob-settings}. The compared baselines are: \noindent\textbf{FLAN-T5-Large} \citep{wei2022finetuned}. This model is fine-tuned on data from over 60 NLP tasks and the authors showed that large-scale instruction tuning significantly improves the model's performance on few-shot reasoning. We evaluate the model's zero-shot performance on temporal reasoning.  \noindent\textbf{ChatGPT} \citep{ouyang2022training}. This model is initialized by GPT-3 and further trained to follow human instructions. We used the \textit{gpt-3.5-0301} version of ChatGPT for more consistent evaluation. Since this model is not open source and not free, we only examined its performance on 200 examples for each setting. \noindent\textbf{T5-SFT} \citep{raffel2020exploring}. This baseline is based on supervised fine-tuning of the conventional T5 models. We use the T5-base model in our experiments and we fine-tune this model on each setting of \tempreason (Section \ref{sec:prob-settings}). 

\begin{table*}
\centering
\resizebox{0.85\textwidth}{!}{
\begin{tabular}{lcccccccccc} 
\hline
                             &                 & \multicolumn{2}{c}{FLAN-T5-L}& \multicolumn{2}{c}{ChatGPT} & \multicolumn{2}{c}{T5-SFT} & \multicolumn{2}{c}{TempT5} &      \\
Question Type               & Setting         & EM   & F1  & EM   & F1                         & EM    & F1                                   & EM    & F1                 & $\Delta$ F1   \\ 
\hline
L1: Time-Time                    & CBQA      & 0.0  & 2.9 & 30.5  & 56.7                        & 100 & 100                                & 100 & 100              & +0.0  \\ 
\hline
\multirow{3}{*}{L2: Time-Event}  & CBQA      & 0.5  & 9.2     & 6.5  & 11.5                    & 1.4   & 23.2                                 & 1.5   & 23.4               & +0.2  \\
                             & ReasonQA & 57.3 & 66.3     & 47.5		 & 51.0                 & 82.6  & 87.1                                 & 84.8  & 88.9               & +1.8  \\
                             & OBQA       & 9.4  & 22.5       & 8.5	 & 16.1                  & 14.8  & 35.2                                 & 15.4  & 36.3               & +1.1  \\ 
\hline
\multirow{3}{*}{L3: Event-Event} & CBQA      & 0.4  & 10.5     & 12.0	  & 21.8                     & 12.1  & 25.3                                 & 12.3  & 25.4               & +0.1  \\
                             & ReasonQA & 36.3 & 47.5    & 49.5	 & 52.3                   & 78.2  & 83.0                                 & 81.1  & 86.1               & +3.1  \\
                             & OBQA       & 8.1  & 19.2       & 17.0	  & 25.3                 & 19.7  & 31.2                                 & 21.1  & 32.4               & +1.2  \\
\hline
\end{tabular}}
\caption{Experimental results of each setting in \tempreason. $\Delta$ F1 refers to the F1 difference between TempT5 and T5-SFT. The reported results are the average scores of three runs.}
\setlength{\belowcaptionskip}{-10pt}
\label{tab:formulation-result}
\end{table*}


\subsection{Experimental Results}
In Table~\ref{tab:formulation-result}, we show the experimental results on the test sets of \tempreason. We then analyze the performance by each level of temporal understanding.

\begin{table}[t]
\centering\resizebox{0.5\columnwidth}{!}{
\begin{tabular}{lcc} 
\hline
          & \multicolumn{2}{c}{TempT5}  \\
Time Range          & EM   & F1                   \\ 
\hline
1000-2022 & 100  & 100                  \\
2022-2040 & 94.4 & 97.1                 \\
\hline
\end{tabular}}

\caption{L1 experimental results of TempT5 on in-domain \tempreason test set and the future test set.}
\label{tab:l1-future}
\end{table}

\noindent\textbf{L1 Understanding}. For L1 temporal understanding, the performance of FLAN-T5-L and ChatGPT significantly deteriorates compared to year prediction (Table~\ref{tab:prelim-exp}). ChatGPT is able to achieve 99.6 EM on year prediction, whereas it can only achieve 30.5 EM on month prediction. The fine-tuned models T5-SFT and TempT5 are able to achieve 100 EM/F1 performance on this task. This showed that even though the L1 logic rules were not explicitly encoded in the language models, we can teach the language model to learn such rules by creating examples of the rules on a large scale. We further evaluate the trained L1-TempT5 model on an out-of-domain futuristic test set (Table~\ref{tab:l1-future}). The questions of the futuristic test set have reference times from 2022 to 2040, which are disjoint from the time period of \tempreason. The TempT5 model performs decently on the future test set, achieving 97.1 F1 score. However, this performance is still below the in-domain performance.   

\noindent\textbf{L2 Understanding}. The time-event relation is the main question type of previous TSQA datasets. When we compare the performance of the three settings of L2 performance, we can see the problem setting plays a significant role. For all three models, the performance of CBQA is the lowest among the three settings. This shows that it is highly difficult for the LMs to answer temporal questions without any context. Meanwhile, ReasonQA has a significantly better performance compared to OBQA and CBQA. This shows that the language models are able to perform temporal reasoning when the relevant facts were provided. That is, once the possible answers and the related timestamps are retrieved, fine-tuned language models (TempT5 and T5-SFT) can perform temporal reasoning relatively well. It is worth noting that the ChatGPT model has the worst performance in the L2 ReasonQA setting while its performance is exceptionally high in the preliminary year prediction experiments. This phenomenon shows that temporal understanding at different levels may not be easily transferable. Last but not least, our proposed TempT5 model achieves significant performance gains over T5-SFT in OBQA and ReasonQA, which is the strongest baseline in our experiments.  

\begin{table}
\centering\resizebox{0.8\columnwidth}{!}{
\begin{tabular}{llcc} 
\hline
                          &  Question Type & EM   & F1    \\ 
\hline
\multirow{2}{*}{L2: CBQA} & P39           & 1.6  & 21.1  \\
                          & Others        & 1.3  & 19.9  \\ 
\hline
\multirow{2}{*}{L3: CBQA} & P39           & 51.4 & 68.2  \\
                          & Others        & 0.6  & 12.1  \\
\hline
\end{tabular}}
\caption{Comparison of L2 and L3 performance of TempT5 in the CBQA setting.}
\label{tab:cbqa-l2l3}
\end{table}

\noindent\textbf{L3 Understanding}. Similar to L2 understanding, all models perform the best in ReasonQA, followed by OBQA and have the worst performance in CBQA. Besides, compared to L2 questions, most models have significantly worse performance on the L3 questions in the ReasonQA setting (except for ChatGPT), showing that L3 temporal reasoning is more challenging than L2. For the FLAN-T5-L model, the performance deterioration from L2 to L3 is 18.8 F1 (L2: 66.3 vs L3: 47.5), whereas the performance gaps of T5-SFT and TempT5 are much lower. It is worth noting that for the T5-SFT model, the exact match scores of L3 questions are significantly higher than those of L2 in the CBQA (L2:1.4 vs L3:12.1) and OBQA (L2:14.8 vs L3:19.7) setting (same for TempT5). We found that this counter-intuitive result is due to a reasoning shortcut of a specific question type `P39 \textit{position held}' (Table~\ref{tab:question-templates}). We further analyze the CBQA performance by question type in Table~\ref{tab:cbqa-l2l3}. For questions other than `P39', L3 performance is significantly worse than L2 (L3: 12.1 F1 vs L2: 19.9 F1). However, the performance of L3 CBQA on `P39' questions is much higher than the other questions. This is because there are reasoning shortcuts for `P39 \textit{position held}' questions from entity names. For example, for the question `Which position did Nicholas Budgen hold before Member of the 46th Parliament of the United Kingdom?', the reasoning shortcut is to simply change the `46th' to `45th'. This shows that L3 temporal reasoning can be achieved via different reasoning paths.

\begin{table}[t!]
\centering\resizebox{0.7\columnwidth}{!}{
\begin{tabular}{lcccc} 
\hline
  & \multicolumn{2}{c}{ReasonQA} & \multicolumn{2}{c}{OBQA}  \\
Metric  & EM   & F1                    & EM   & F1                 \\ 
\hline
TempT5   & 84.8 & 88.9                  & 15.4 & 36.3               \\
--TSE    & 84.0 & 88.0                  & 14.8 & 35.5               \\
--TSRL   & 83.4 & 87.7                  & 15.0 & 35.8               \\
\hline
\end{tabular}}
\setlength{\belowcaptionskip}{-10pt}
\caption{Ablation analysis of TempT5 based on L2 questions.}
\label{tab:ablation}
\end{table}

\begin{table}[t!]
\centering
\resizebox{\columnwidth}{!}{
\begin{tabular}{ccccc} 
\hline
 &  & FLAN-T5-L & ChatGPT & TempT5  \\ 
Time Range & \% Train
    & F1    & F1      & F1 \\ 
\hline
before 1900       &    8.4       & 69.5  & 77.8  & 85.6 \\
1900-1920        &   4.1  & 67.9   & \textbf{78.7}  & 87.5  \\
1920-1940        & 6.6  & 65.3  & 43.8  & 87.6 \\
1940-1960        &  7.5   & \textbf{71.9}   & 47.9   & 88.7 \\
1960-1980        &   11.0  & 68.0 & 43.8  & \textbf{90.5}  \\
1980-2000        &    18.3  & 65.6  & 43.9  & 89.6  \\
2000-2020        &  37.8   & 66.1   & 49.1  & 89.8 \\
2020-2040        &   6.3  & 68.5   & 72.7  & 82.6 \\
Overall         &  100  & 67.1   & 51.0  & 88.9 \\
\hline
\end{tabular}}
\caption{Performance breakdown of different models in L2 ReasonQA across different time periods. We can see that ChatGPT has the worst performance among the three models and its performance is highly variable across different time periods.}
\label{tab:l2-reasoning-by-year}
\end{table}

\subsection{Ablation Study}
In Table~\ref{tab:ablation}, we showed the ablation study of TempT5 based on the L2 questions in the OBQA and ReasonQA settings. We can see that TSE and TSRL have different effects in the two settings. Removing TSRL has a heavier impact on the ReasonQA setting, leading to a 1.2 F1 drop. On the other hand, TSE pre-training is more important in the OBQA setting and removing the TSE pre-training leads to a performance drop of 0.8 F1.

\begin{table}[t!]
\centering
\resizebox{0.9
\columnwidth}{!}{
\begin{tabular}{llcc} 
\hline
                              & Question Type & EM   & F1    \\
\hline
\multirow{2}{*}{L2: ReasonQA} & Intra-year    & 80.5 & 86.3  \\
                              & Inter-year    & 86.9 & 90.3  \\
\hline
\end{tabular}}
\caption{Performance of TempT5 in L2 ReasonQA by question type. The intra-year question type refers to questions that have multiple possible answers within one year. In contrast, the inter-year question type only has one possible answer in that specific year.}
\label{tab:intra-year}
\end{table}

\begin{table}[t]
\centering
        \resizebox{1\columnwidth}{!}{
        \begin{tabular}{p{10cm}}
            \toprule
            \textit{\textbf{Example 1}}\quad\quad\textit{\textbf{Error Type: Intra Year Error}}\\
            \textbf{Error Cause}: Lack of monthly-level understanding. 
            
            \textbf{Question}: Which position did Hirofumi Yoshimura hold in Jul 2019?
            \quad\quad

             \textbf{Context}: 
            Hirofumi Yoshimura holds the position of: 
            
            Governor of Osaka Prefecture from Apr 2019 to Dec 2022. 
            
            Member of the House of Representatives of Japan from Dec 2014 to Oct 2015. 
            
            Mayor of Osaka from Dec 2015 to Mar 2019.

             \\
             \textbf{Prediction}: Mayor of Osaka 
             \\
             \textbf{Ground Truth}: Governor of Osaka Prefecture
             \\
             \bottomrule
        \end{tabular}}

    \caption{An example of intra-year error of TempT5 in L2 ReasonQA.}
    \label{tab:error-analysis}
    \end{table} 
\subsection{Further Analysis}
\label{sec:analysis}
In this section, we examine the model biases in \tempreason. We first analyze the L2 reasoning performance across different years in a similar manner as Section \ref{sec:prelim}. The performance breakdown can be found in Table~\ref{tab:l2-reasoning-by-year}. We can see that for the FLAN-T5-L model and ChatGPT model, the L2 reasoning performance fluctuates across different time periods. FLAN-T5-L not only has higher performance but also lower variability across the different time periods. On the other hand, from the performance breakdown of our proposed TempT5, we can see that the temporal biases shown in the year prediction experiments (Table~\ref{tab:prelim-exp}) were alleviated. The F1 scores from 1940 to 2020 were similar. However, the F1 scores before 1900 and after 2020 are still significantly worse than the other time periods. This performance degradation is largely due to the lack of training data in those time periods.  

The other major source of errors comes from the intra-year question type. The intra-year question type refers to questions that have multiple possible answers within one year. Therefore, it requires reasoning at the month level. As shown in Table~\ref{tab:intra-year}, the performance on intra-year questions is significantly worse than the performance on inter-year questions, especially for the difference in the EM score (6.4, 86.9 vs. 80.5). In Table~\ref{tab:error-analysis}, we show an example of an intra-year reasoning error. We can see that the model fails to capture the intra-year position change of the subject.

\section{Related Work}
\paragraph{Temporal Information Extraction} Early efforts on temporal NLP research primarily focus on event temporal information extraction. \citet{pustejovsky2003timebank} constructed the TimeBank corpus, which is a temporally annotated corpus that annotates events, times, and temporal relations (such as before/after). The TIE task asks models to extract the events within a piece of text and to identify the temporal relations between event pairs. The TempEval (\citealp{verhagen-etal-2010-semeval}; \citealp{bethard-etal-2016-semeval}) challenge is a popular TIE challenge with a similar annotation scheme as TimeBank. However, it is costly to exhaustively annotate the temporal relations among all events. \citet{cassidy-etal-2014-annotation} proposed a dense annotation scheme and constructed the TimeBank-Dense dataset, which has more complete annotation compared to TimeBank. \citet{han-etal-2019-joint} proposed a joint framework to extract events and time in an end-to-end manner. \citet{rajaby-faghihi-kordjamshidi-2021-time} proposed the Time-stamped Language Model to understand the flow of events. However, prior works in this field focused on extracting events and temporal relations within one piece of document. The models trained on this task cannot perform global event-to-time grounding.

\paragraph{Temporal Reasoning over KGs} The Temporal Knowledge Graph Completion (TKGC) field studies temporal reasoning in knowledge graphs. This task aims to rank all entities in a knowledge graph given a temporal query. Many works in this field (TTransE, \citealp{jiang-etal-2016-towards}; TTransH, \citealp{dasgupta-etal-2018-hyte}; TNTComplEx, \citealp{Lacroix2020Tensor}) were proposed as extensions to prior knowledge completion techniques, such as TransE \citep{bordes2013transe}, TransH \citep{Wang_Zhang_Feng_Chen_2014}, and ComplEx \citep{kipf2017semisupervised}. With a similar concept as TKGC, several question answering datasets are proposed based on temporal knowledge graphs, such as TEQUILA \citep{jia2018tequila}, TimeQuestions \citep{jia2021complex}, and CronQuesions \citep{saxena-etal-2021-question}. These datasets include more complex questions in a natural language format, and the task setting is also asking models to rank all the entities of a given knowledge graph. \citet{mavromatis2022tempoqr} proposed a joint model that unifies temporal KG embeddings and pre-trained language models for this task. \citet{shang-etal-2022-improving} proposed a contrastive approach to improve the QA performance for temporal KGs. Temporal reasoning in KGs is closely related to our problem of interest. However, the major difference is that KGQA presumes all the entities are known to the system and the task is to rank all the possible entities that satisfy the queries. In contrast, our task aims to answer temporal questions based on natural text input only.

\paragraph{Temporal Reasoning for LMs} Large language models (\citealp{devlin-etal-2019-bert}; \citealp{raffel2020exploring}; \citealp{liu2019roberta}) have demonstrated good performance on the question answering task (\citealp{rajpurkar-etal-2016-squad}; \citealp{kwiatkowski2019natural}). In recent years, several contemporary time-sensitive QA datasets were proposed. \citet{zhang-choi-2021-situatedqa} proposed the SituatedQA dataset, which contains plenty of time-dependent question-answer pairs. The \templama dataset \citep{10.1162/tacl_a_00459} was proposed to evaluate the CBQA performance for time-dependent questions from 2010 to 2020. However, the QA performance of \templama may be overestimated, since it only covers a short time period and the period is in favor of LM's temporal bias. Similarly, StreamingQA \citep{liska2022streamingqa} has a similar disadvantage, since its time coverage is from 2007 to 2020. The Time-sensitive QA dataset \citep{chen2021dataset} covers a relatively longer timespan (from 1367 to 2018), but it only contains questions of time-event relation. The common drawback of the previously proposed TSQA datasets is the lack of coverage of temporal reasoning levels other than the time-event type of questions.

\section{Conclusions and Future Work}
In this paper, we tackled the under-explored temporal reasoning problem for large language models. We found that large language models are highly susceptible to biases of time, and their temporal reasoning capability varies depending on the specific time given in the question. Besides, we proposed a comprehensive time-sensitive QA dataset \tempreason to evaluate LMs' temporal reasoning capability in diverse settings. Lastly, we proposed a novel training paradigm to improve language models' reasoning capability by temporal span extraction pre-training and time-sensitive reinforcement learning. We conducted extensive experiments and demonstrated that our proposed model consistently outperformed strong baselines.

\section{Limitations}
\label{sec:limitation}
The focus of the \tempreason dataset is to examine language models' temporal reasoning capability. However, the temporal expressions of \tempreason are only in the form of month in textual form and year in numeric form. One limitation of the \tempreason benchmark is the lack of adversarial attacks in other temporal formats, such as all numeric dates and months. The robustness of temporal reasoning is also important in real-world applications. Since the scope of this paper only focuses on the reasoning aspect, the robustness of \tempreason will be left for future research. Besides, the knowledge triples of \tempreason are from the crowd-sourced Wikidata KB, and these triples are used to construct the question-answer pairs in this paper. Hence, it is possible that errors in the Wikidata KB propagate to the answers in \tempreason. However, such errors have minimal effect in the ReasonQA setting, for this task only asks the models to infer from factual knowledge in the Wikidata KB.

\section{Ethics Statement}
\label{sec:ethics}
In this paper, we created a probing dataset \tempreason for temporal reasoning evaluation. The dataset is constructed based on the matching of Wikidata KB and Wikipedia articles. This approach is commonly used for distantly supervised data construction. The Wikidata KB is under the public domain\footnote{\url{https://www.wikidata.org/wiki/Wikidata:Licensing}} and the Wikipedia articles are licensed under the Creative Commons AttributionShareAlike 3.0 Unported License\footnote{\url{https://en.wikipedia.org/wiki/Wikipedia:Copyrights}}. Therefore, we are able to adapt these data to construct our dataset. We will also release our data under the same license as Wikidata. The scope of our dataset is purely for scientific research of language models' temporal reasoning capability. However, the contexts from the Wikipedia articles may contain improper content. The adoption of such content is not a decision of the authors, and all content in the dataset does not reflect
the views or stances of the authors of this paper.

\section{Acknowledgements}
We would like to thank all the reviewers for their insightful comments and constructive feedback.

\bibliography{custom}
\bibliographystyle{acl_natbib}

\appendix
\begin{table*}[ht]
\resizebox{\textwidth}{!}{
\begin{tabular}{lllllllll}
\toprule
Dataset           & QA format & Knowledge Corpus           & Closed/Open/Reason & Time Coverage & Size          & L1 & L2 & L3 \\
\midrule
\tempreason        & Language  & Wikidata/Wikipedia         & Closed/Open/Reason & 634-2023      & 52.8K  &  \checkmark  &  \checkmark  & \checkmark   \\
\templama \citep{10.1162/tacl_a_00459}         & Language  & Wikidata                   & Closed        & 2010-2020     & 50k           & \xmark   &  \checkmark    & \xmark   \\
Time-Sensitive QA \cite{chen2021dataset} & Language  & Wikidata/Wikipedia         & Open               & 1367-2018     & 41.2k         & \xmark   &  \checkmark  &  \xmark  \\
StreamingQA \citep{liska2022streamingqa}       & Language  & WMT                        & Closed/Open        & 2007-2020     & 147k          & \xmark   & \checkmark   & \xmark   \\
SituatedQA \citep{zhang-choi-2021-situatedqa}       & Language  & Wikipedia/Human Annotation & Closed/Open        & 1270-2021     & 12.2k         & \xmark   &  \checkmark  & \xmark  \\
\midrule
TempQuestions \citep{jia2018temp}       & KG  & Freebase & KG        & NA     & 1.2k         & \xmark   &  \checkmark  & \checkmark  \\
TimeQuestions \citep{jia2021complex}       & KG  & Wikidata & KG        & NA     & 16.1k         & \xmark   &  \checkmark  & \checkmark  \\
CronQuestions \citep{saxena-etal-2021-question}       & KG  & Wikidata & KG        & 34-2021    & 410k         & \xmark   &  \checkmark  & \checkmark  \\
\bottomrule
\end{tabular}}
\caption{Dataset comparison of \tempreason and prior datasets.}
\label{tab:dataset-comparison}
\end{table*}

\section{Realtime Adaptation of LMs}
\label{sec:realtime-adapt}

Besides the experiments on our proposed \tempreason dataset. We also examined our model in the RealtimeQA \citep{kasai2022realtime} leaderboard. This leaderboard releases time-sensitive questions every week based on weekly quizzes from news websites (such as CNN and CNBC). The RealtimeQA challenge has two tracks: (1) multiple-choice questions and (2) generation track. The generation track of this challenge is the same as OBQA in this paper. We examined our model along with the two retrievers provided in the challenge: (1) Google custom search (GCS), and (2) Dense Passage Retrieval (DPR, \citealp{karpukhin-etal-2020-dense}). We adapt our TempT5 model of L2 ReasonQA on the question-answer pairs of RealtimeQA before December 2022. We then evaluate the adapted model on the questions released on 16th December 2022\footnote{\url{https://realtimeqa.github.io/docs/results/2022/20221216}}. Experimental results (Table~\ref{tab:realtimeqa}) show that our model performs competitively even when adapting to the most up-to-date TSQA challenge.

\begin{table}[ht]
\centering
\begin{tabular}{lcc} 
\hline
           & EM   & F1    \\ 
\hline
GPT3+GCS$^\dagger$   & 55.0   & 63.6  \\
TempT5-L+GCS & 48.3 & 53.3  \\
RAG+GCS$^\dagger$    & 35.0   & 45.9  \\
GPT3+DPR$^\dagger$    & 17.2   & 23.0  \\
TempT5-L+DPR & 10.3 & 18.4  \\
RAG+DPR$^\dagger$    & 0.0   & 3.1  \\
\hline

\end{tabular}

\caption{Experimental results on the generation track of RealtimeQA leaderboard based on December 16, 2022‘s questions. The task formulation of this track is the same as OBQA in this paper. Results with $\dagger$ are taken from the URL in the footnote.}
\label{tab:realtimeqa}
\end{table}

\label{sec:question-templates}
\begin{table*}[t]
\small
\centering
\resizebox{0.8\textwidth}{!}{
\begin{tabular}{@{}clcl@{}}
\toprule
\textbf{WikiData ID} & \textbf{KB Relation}              & \textbf{\# Queries} & \textbf{Template}   
 \\ \midrule
\textit{\textbf{L1 Question Templates:}}&   &        &                      \\ \midrule
NA         & NA &   NA     &  What is the time $x$ year(s) and $y$ month(s) before/after $t$?  \\ 
NA         & NA &   NA     &  What is the time $x$ year(s)  before/after $t$?  \\ 
NA         & NA &   NA     &  What is the time $y$ month(s)  before/after $t$?  \\ \midrule
\textit{\textbf{L2 Question Templates:}}&   &        &                      \\ \midrule
P54         & member of sports team & 4,087       & Which team did \textless{}subject\textgreater~ play for in $t$?                       \\
P39         & position held         & 3,133       & Which position did \textless{}subject\textgreater~ hold in $t$?            \\
P108        & employer              & 2,368       & Which employer did \textless{}subject\textgreater~ work for in $t$?.                        \\
P102        & political party       & 500       & Which political party did \textless{}subject\textgreater~  belong to in $t$?               \\
P286        & head coach            & 1,153       & Who was the head coach of \textless{}subject\textgreater{} in $t$?             \\
P69         & educated at           & 750       & Which school was  \textless{}subject\textgreater{} attending in $t$?                         \\
P488        & chairperson           & 1,904       & Who was the chair of \textless{}subject\textgreater{} in $t$?                 \\
P6          & head of government    & 1,627       & Who was the head of the government of \textless{}subject\textgreater{} in $t$? \\
P35          & head of state    & 250       & Who was the head of the state of \textless{}subject\textgreater{} in $t$? \\
P127        & owned by              & 245       & Who was the owner of \textless{}subject\textgreater in $t$?                  \\ \midrule
\textit{\textbf{L3 Question Templates:}}&   &        &                      \\ \midrule
P54         & member of sports team & 2,524       & Which team did \textless{}subject\textgreater~ play for before/after $o_{j}$?                       \\
P39         & position held         & 2,538       & Which position did \textless{}subject\textgreater~ hold before/after $o_{j}$?            \\
P108        & employer              & 1,991       & Which employer did \textless{}subject\textgreater~ work for before/after $o_{j}$?.                        \\
P102        & political party       & 433       & Which political party did \textless{}subject\textgreater~  belong to before/after $o_{j}$?               \\
P286        & head coach            & 1,051       & Who was the head coach of \textless{}subject\textgreater~ before/after $o_{j}$?             \\
P69         & educated at           & 594       & Which school was  \textless{}subject\textgreater~ attending before/after $o_{j}$?                         \\
P488        & chairperson           & 1,881       & Who was the chair of \textless{}subject\textgreater~ before/after $o_{j}$?                 \\
P6          & head of government    & 1,535       & Who was the head of the government of \textless{}subject\textgreater~ before/after $o_{j}$? \\
P35          & head of state    & 268       & Who was the head of the state of \textless{}subject\textgreater~ before/after $o_{j}$? \\
P127        & owned by              & 199       & Who was the owner of \textless{}subject\textgreater~ before/after $o_{j}$?                  \\ \midrule

\end{tabular}}
\caption{Templates used for converting Wikidata facts into natural questions. For the L2 questions, $t$ is a randomly sampled time between the start time $t_{s}$ and end time $t_{e}$ of the given fact. The format of $t$ is month and year (examples shown in Figure \ref{fig:temp-r-example}). $o_{j}$ refers to the object entity name that is before or after the correct answer. The numbers of queries are from the L2 and L3 training sets of \tempreason.}
\label{tab:question-templates}
\end{table*}

\section{Implementation Details}
\label{sec:implementation}
This section describes the implementation details of our models and baselines. For temporal span extraction pre-training, we use the T5-base model for initialization. We then train the model for 100K steps with a batch size of 8 and a learning rate of 2e-5. We use the maximum input length of 512 for TSE pre-training. For task-specific fine-tuning, we use the same batch size and learning rate, whereas the maximum input lengths are different for each setting. For the CBQA setting, the maximum input length is set as 128, since only the question is given to the model. For the ReasonQA setting, the maximum input length is set as 512. The maximum length of 1,024 is used for the OBQA setting, since the context in this setting is the longest on average. For each setting, we fine-tune the language model for 3 epochs, and evaluation is conducted using the final checkpoint. For time-sensitive reinforcement learning, we followed the proximal policy optimization (PPO, \citealp{schulman2017proximal}) algorithm. Instead of using a reward model, we use the reward function described in Section~\ref{sec:method}. For this stage, we set the initial KL penalty coefficient as 0.05 and the target KL coefficient as 6. The discount factor $\gamma$ for PPO is set to 0.99.

\section{Comparison of \tempreason and Prior Datasets}
\label{sec:compare-dataset}

In Table~\ref{tab:dataset-comparison}, we show the detailed comparison of our \tempreason dataset and prior time-sensitive question answering datasets. Our dataset is the first to include all three temporal reasoning types and the ReasonQA setting.

\section{\tempreason Templates}
\label{sec:templates}
The templates that we used to create \tempreason is shown in Table~\ref{tab:question-templates}.

\end{document}